\pdfoutput=1
\documentclass{llncs}
\usepackage[english]{babel}
\usepackage[utf8]{inputenc}
\usepackage{graphicx}
\usepackage[percent]{overpic}
\usepackage[noabbrev,capitalise]{cleveref}
\usepackage{xcolor}
\usepackage{amssymb}
\usepackage{amsmath}
\usepackage{subcaption}

\newcommand\Tstrut{\rule{0pt}{2.6ex}}         
\newcommand\Bstrut{\rule[-0.8ex]{0pt}{0pt}}   

\begin{document}

\title{Combining Stereo Disparity and Optical Flow for Basic Scene Flow}
\author{René Schuster, Christian Bailer, Oliver Wasenmüller, Didier Stricker}
\institute{DFKI -- German Research Center for Artificial Intelligence\\
\email{firstname.lastname@dfki.de}}

\maketitle

\begin{abstract}
Scene flow is a description of real world motion in 3D that contains more information than optical flow. Because of its complexity there exists no applicable variant for real-time scene flow estimation in an automotive or commercial vehicle context that is sufficiently robust and accurate. Therefore, many applications estimate the 2D optical flow instead. In this paper, we examine the combination of top-performing state-of-the-art optical flow and stereo disparity algorithms in order to achieve a basic scene flow. On the public KITTI Scene Flow Benchmark we demonstrate the reasonable accuracy of the combination approach and show its speed in computation.
\end{abstract}

\section{Introduction} \label{sec:intro}
Development and use of Advanced Driver Assistance Systems (ADAS) have become a more and more relevant topic in automotive applications. This is not only concerning passenger cars, but also commercial vehicles in agriculture or transportation. Increased usability, efficiency, and safety is of high importance for the driver, the manufacturer, and other road users.
Typical high-level tasks comprise steering and speed assistance, driver monitoring, and early-warning systems that all rely on an accurate perception and recognition of the environment. A detailed reconstruction of the 3D geometry of the surroundings as well as precise estimation of the motion of other traffic participants are the core components of this reconstruction. Despite the fast progress in depth estimation and 2D optical flow computation, the real-world representation of motion in 3D -- scene flow -- has not yet found its way into serial automotive applications. This might be due to the increased complexity of the problem. Therefore, motion perception in vehicles is often approximated by optical flow. Yet, especially applications for agricultural vehicles could benefit greatly from a detailed estimation of 3D motion, e.g. to avoid collisions with animals. In this scenario, a scene flow based motion estimator could also detect animals that move in the direction of viewing, while vision based motion estimation in 2D can only detect animals that move vertically to the direction of driving.

\begin{figure}[ht]
\centering
\includegraphics[width=\textwidth]{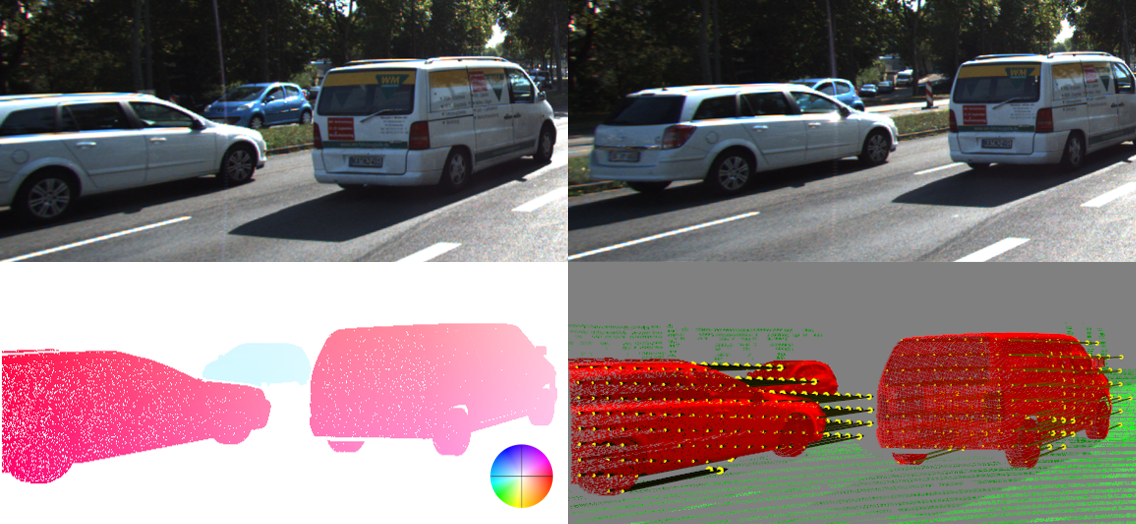}
\caption{Comparison of Optical Flow and Scene Flow. For four stereo images of a traffic scene (top row showing the two images of the left camera), the optical flow (bottom left) displays motion parallel to the image plane only. Scene flow (bottom right) -- here visualized with 3D vectors in a point cloud -- gives full 3D motion information and additionally reconstructs the full 3D geometry.}
\label{fig:ofvssf}
\end{figure}

However, 3D scene flow can be computed by the combination of 2D optical flow and depth from stereo disparity. In this paper, we use this combination approach to compute basic scene flow and show its advantages over 2D optical flow. Further, we demonstrate its reasonable accuracy on the KITTI Scene Flow Benchmark \cite{menze2015object}, where we outperform many dedicated scene flow algorithms.

\section{Related Work} \label{sec:related}
The estimation of scene flow in traffic scenarios experienced remarkable improvements by the publication of the first scene flow benchmark consisting of realistic image data, the KITTI Scene Flow Benchmark \cite{menze2015object}.
Before, most methods were only able to compute scene flow with reasonable high precision in a controlled indoor environment. Variational approaches, like \cite{vedula1999three,huguet2007variational,basha2013multi}, still are not applicable outdoors in an automotive context where fast motions, large distances, and many objects are observed.
Vogel et. al \cite{vogel2013PRSF} have introduced the piece-wise rigid scene flow that allowed for a very strong spatial regularization. This model has later been adopted by \cite{menze2015object,lv2016CSF} with changes to the optimization procedure. Multi-frame extensions \cite{neoral2017object,vogel2015PRSM} have shown to further improve accuracy by enforcing consistency over time. Though all theses approaches brought great progress for scene flow estimation in traffic scenarios, they all have typically very long run times due to the complexity of the optimization. Making them inappropriate for use in any application that has time constraints. Contrary to that, our proposed method has a potentially short run time which is real-time capable.
Recent methods that avoid strong regularization or a piece-wise planar motion model \cite{taniai2017fsf,schuster2018sceneflowfields} have shown to be considerable faster while maintaining a state-of-the art performance on different data sets. Still, the run time is far from real-time.

Others have tried to compute scene flow from a combination of stereo depth and 2D optical flow, but back then optical flow estimation was not as advanced as it is nowadays.
The pre-computation of depth to generate input for methods that require RGB-D images for scene flow estimation could be considered a weakened form of our proposed combination approach, e.g \cite{hornacek2014sphereflow,jaimez2015primal}. Though, a robust depth estimation with active cameras is still a challenging task, especially in an automotive context \cite{yoshida2017time}.

Stereo and optical flow algorithms have a longer history \cite{fortun2015optical}. The amounts of different approaches, data sets, and benchmarks are bigger compared to the more complex scene flow problem. Therefore, depth estimation from stereo images and optical flow computation do not belong to the remaining challenges. Rather, for special setups, both tasks can be considered as solved.
Since the recent rise of artificial neural networks, many methods that apply deep learning have been proposed. However, large neural networks often do not only increase the accuracy but also the run time. 
Two very popular stereo algorithms are \cite{hirschmuller2008SGM,yamaguchi2014efficient} because they have achieved very good accuracy and speed at the same time. That's why they are often used as standalone algorithms or for initialization purposes.
For optical flow, we like to highlight the Flow Fields family. The first approach was published in \cite{bailer2015flow} and has since then been steadily refined to improve accuracy. The latest version even uses deep learning to match correspondences across the images \cite{bailer2017cnn}. These methods are noteworthy because they were among the first that achieved top performance across different data sets which makes them very versatile.
For our combination approach, we use Semi-Global Matching (SGM) \cite{hirschmuller2008SGM} and Flow Fields+ \cite{bailer2017optical}, which both have state-of-the-art accuracy while at the same time being reasonably fast. 

\section{Scene Flow from Stereo and Optical Flow} \label{sec:method}

\begin{figure}[ht]
	\centering
	\begin{overpic}[width=0.48\textwidth]{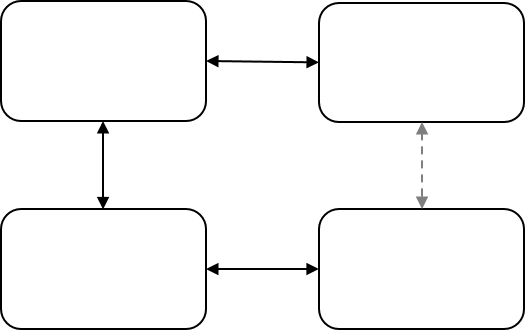}
		\put(16,8){\Large$I_{l}^{t}$}
		\put(77,8){\Large$I_{r}^{t}$}
		\put(14,48){\Large$I_{l}^{t+1}$}
		\put(75,48){\Large$I_{r}^{t+1}$}
		\put(47,13){\large$d^t$}
		\put(44,53){\large$d^{t+1}$}
		\put(12,28){\large$\textbf{u}_l$}
		\put(72,28){\large\color{darkgray}$\textbf{u}_r$}
	\end{overpic}
	\caption{Relation of two stereo image pairs. Subscripts denote the view point (left/right) and superscripts the time step. We consider the left image at time $t=0$ the reference frame. Each stereo image pair is related by the according disparity map, while the temporal image pairs are related by 2D optical flow.} \label{fig:setup}
\end{figure}
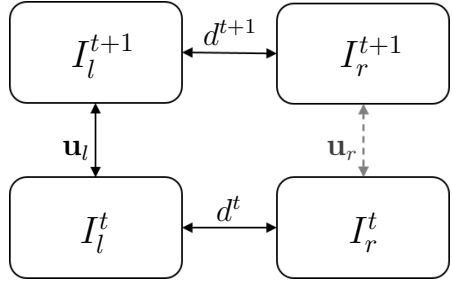

As usual for scene flow algorithms, we assume a standard stereo camera rig consisting of calibrated left and right cameras that is built-in in most passenger cars and commercial vehicles. As input for our approach, two rectified temporally adjacent frame pairs are used (see \cref{fig:setup}) that share the same image domain $\Omega$, i.e. they have the same size. These four images provide sufficient information to estimate 3D scene flow.

To avoid greater overhead in computation and still estimate a full scene flow representation in 3D, we want to combine depth and 2D motion information. Depth is estimated by stereo disparity using only a single image pair of the left and right camera at a time. Motion information is estimated by optical flow using only consecutive images from the left camera over time. Both complement each other regarding scene flow. Optical flow is lacking of depth information, and stereo depth can be generalized to change over time. An example of the two images of the left camera, the computed optical flow, and the reconstructed scene flow is given in \cref{fig:ofvssf}. It is evident that the scene flow representation of the displayed scenario contains more information about the real world motion.
In the following, we will give some details about the stereo and optical flow algorithms which we will use to reconstruct scene flow. The combination itself is described in \cref{sec:combination}.

\subsection{Depth Estimation by Semi-Global Matching} \label{sec:sgm}
Depth estimation wants to recover the shortest distance between the observed scene and the image plane for each pixel. This is usually done by triangulation based on different view points that capture the same scene. The standard case considers only two different views, a left and a right view, like in the human visual system. For simplification, the two images get rectified so that both images lie on the same plane and are horizontally aligned. This way, the corresponding image points in the two images share the same horizontal line which reduces the search space from two to one dimension. A stereo algorithm tries to find the best correspondences for all pixels of one image so that a consistent depth map is created. To this end a regularization is often applied that enforces smoothness. The goal is to find a mapping that maps each pixel to its depth value. Since the real depth and the disparity, i.e. the offset between the corresponding left and right image points, are directly related by the projection of the camera, both representations are equivalent if the camera parameters are known. For our work, we consider a disparity map
\begin{equation} \label{eq:depth}
\mathcal{D}: \Omega \rightarrow \mathbb{R}^+
\end{equation}

Stereo matching in SGM \cite{hirschmuller2008SGM} is not using a local nor global neighborhood for regularization. Instead a semi-global energy formulation which uses eight different paths radiating from each target pixel location is used which enables sharp boundaries, accurate depth estimation and good run time. 
Because of occlusions, it is hard to recover depth for all image points. Yet it is possible to detect potentially wrong estimates through consistency checks. SGM uses a left-right consistency check for two computed depth maps to localize and remove wrongly estimated values. Thus, SGM yields a non-dense disparity map.

\subsection{Optical Flow from Flow Fields+} \label{sec:ff+}
Optical flow is the estimation of 2D motion in a sequence of at least two images. Each pixel gets mapped to a 2D vector in image space that indicates the corresponding pixel in the next time step of the image sequence.
\begin{equation} \label{eq:flow}
\mathcal{F}: \Omega \rightarrow \mathbb{R}^2
\end{equation}

The problem is related to stereo algorithms in that it also tries to find matches across two images. But for optical flow, the search space is much bigger as there is no epipolar constraint that restricts the matching to a linear search.

Flow Fields \cite{bailer2015flow} tackles this problem without any regularization which makes it so versatile. It is tailored to find pixel correspondences for optical flow estimation by propagation and random search with multiple stages of outlier filtering followed by interpolation with EpicFlow \cite{revaud2015epic} to reconstruct an accurate, dense optical flow field.
Flow Fields+ \cite{bailer2017optical} is the extension of Flow Fields \cite{bailer2015flow} that is using an enhanced matching term.
In detail, Flow Fields+ is initialized on a lower resolution by matching Walsh-Hadamard features \cite{hel2005real} using k-dimensional search trees. For several iterations, these initial flow values get propagated into their local neighborhood. After each iteration, a random search is performed that allows to refine the propagated values. The fitness of a match is determined by a patch-based data term. After all iterations, the estimated flow field gets lifted to the next higher resolution until full resolution is reached.
Similar as in the stereo algorithm, the full resolution optical flow can contain wrong estimates that needs to be filtered. To this end, Flow Fields+ computes two additional inverse flow fields. If any match is inconsistent with any of the inverse fields, it gets filtered. EpicFlow \cite{revaud2015epic} is used to fill up the gaps. EpicFlow is an edge-aware interpolation method for optical flow, that is used as post-processing step after filtering in many optical flow methods. Finally, a dense optical flow map is obtained.

\subsection{The Combination Approach} \label{sec:combination}

Using SGM and Flow Fields+, we compute the depth maps $\mathcal{D}^0$ at time $t=0$ and $\mathcal{D}^1$ at time $t=1$ and the flow map $\mathcal{F}$ (cf. \cref{fig:setup}). From these three mappings, it is possible to reconstruct a scene flow field that we define as follows:
\begin{equation}
\mathcal{S}: \Omega \rightarrow \mathbb{R}^4
\end{equation}

Each image point $\mathbf{x} = (x,y)^T$ is mapped to four values $\mathcal{S}(\mathbf{x}) = (u,v,d_0,d_1)^T$ which fully describe the 3D motion and 3D geometry given the camera intrinsics and extrinsics. The four components are the values of the optical flow field, the disparity at the current time step, and the disparity value at the next time step where the optical flow is pointing to.
The combination method is straightforward with almost no additional computation time and can formally be described as follows.
\begin{equation}
\mathcal{S}(\mathbf{x}) = 
	\begin{cases}
	(\mathcal{F}(\mathbf{x})^T, \mathcal{D}^0(\mathbf{x}), \mathcal{D}^1(\mathbf{x}+\mathcal{F}(\mathbf{x})))^T, & \text{if}\ \mathbf{x},\mathbf{x}+\mathcal{F}(\mathbf{x}) \in \Omega\ \text{and}\ \\
	& \mathcal{D}^0(\mathbf{x}), \mathcal{D}^1(\mathbf{x}+\mathcal{F}(\mathbf{x}))\ \text{valid}\\
	\text{undefined}, & \text{otherwise} 
	\end{cases}
\end{equation}

We warp the disparity map for the temporally second image pair according to the estimated optical flow. The values $\mathcal{D}^1(\mathbf{x}+\mathcal{F}(\mathbf{x}))$ at sub-pixel positions get interpolated using bilinear interpolation:
\begin{equation}
	\begin{split}
	\mathcal{D}(x',y') =\ & \mathcal{D}(\lfloor x' \rfloor,\lfloor y' \rfloor) \cdot (1-\{x'\}) \cdot (1-\{y'\}) \\ 
	&+ \mathcal{D}(\lfloor x' \rfloor + 1,\lfloor y' \rfloor) \cdot \{x'\} \cdot (1-\{y'\}) \\
	&+ \mathcal{D}(\lfloor x' \rfloor,\lfloor y' \rfloor +1) \cdot (1-\{x'\}) \cdot \{y'\} \\
	&+ \mathcal{D}(\lfloor x' \rfloor +1,\lfloor y' \rfloor +1) \cdot \{x'\} \cdot \{y'\}
\end{split}
\end{equation}

Where $\mathbf{x}' = \mathbf{x}+\mathbf{u} = \mathbf{x} + \mathcal{F}(\mathbf{x})$ is the target position of pixel $\mathbf{x}$, and $\{x\} = x - \lfloor x \rfloor$ denotes the fractional part of number $x$.
The difference of the disparity values yields the missing motion component in direction of viewing. 

However, if the optical flow leaves the image boundaries, or the filtered depth maps contain gaps, we can not reconstruct the full scene flow at every pixel position. Thus, our scene flow estimate is non-dense by nature. This is also reflected in the results shown in \cref{tab:kitti} where the evaluation results for our original estimated sparse scene flow (\textit{Est}) and for a dense version (\textit{All}) that was interpolated by KITTI during evaluation are presented.

\section{Results} \label{sec:results}
\begin{figure}[p]
	\begin{center}
		\begin{subfigure}[c]{1\textwidth}
			\centering
			\includegraphics[width=0.44\textwidth]{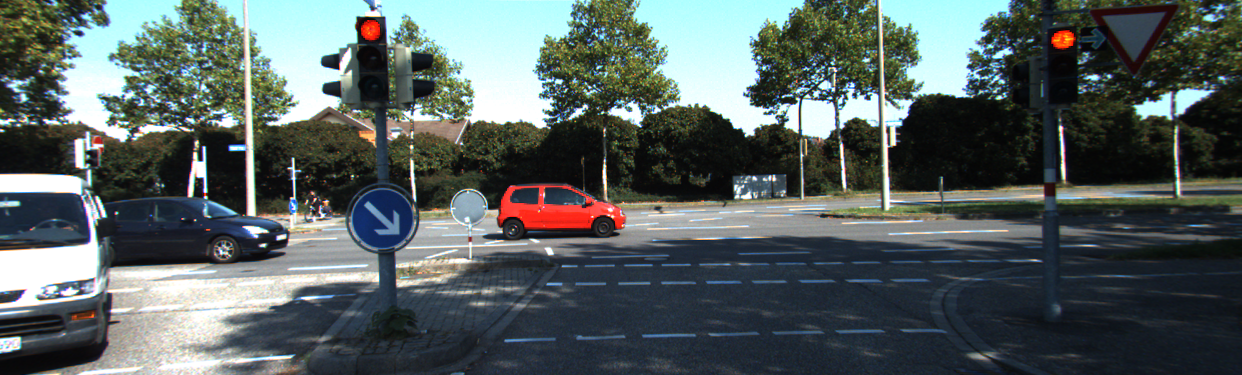}%
			\vspace{1mm}%
		\end{subfigure}
		\hfill
		\begin{subfigure}[c]{0.44\textwidth}
			\centering
			Estimates%
			\vspace{0.5mm}%
			\end{subfigure}
		\begin{subfigure}[c]{0.44\textwidth}
			\centering
			Error Maps%
			\vspace{0.5mm}%
		\end{subfigure}
		\hfill%
		\\%
		\hspace{0.04\textwidth}
		\begin{subfigure}[c]{0.44\textwidth}
			\includegraphics[width=1\textwidth]{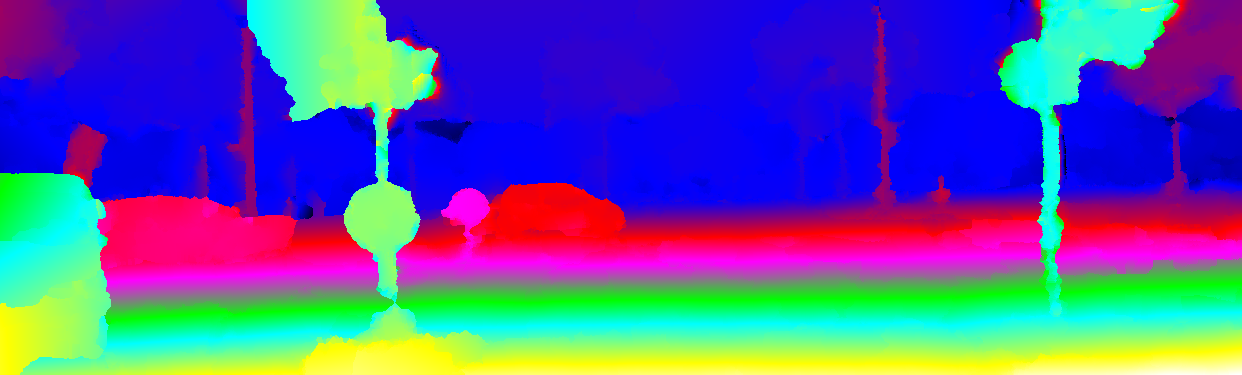}%
			\vspace{0.5mm}%
		\end{subfigure}
		\begin{subfigure}[c]{0.44\textwidth}
			\includegraphics[width=1\textwidth]{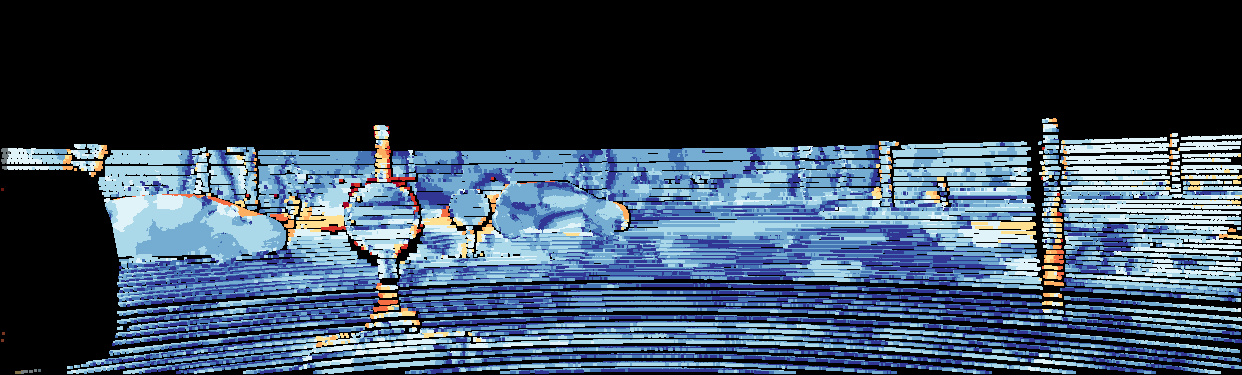}%
			\vspace{0.5mm}%
		\end{subfigure}
		\begin{subfigure}[c]{0.04\textwidth}
			\rotatebox[origin=c]{90}{D1}%
		\end{subfigure}
		\\%
		\begin{subfigure}[c][5mm][c]{0.04\textwidth}
			\rotatebox[origin=c]{90}{SceneFlowFields \cite{schuster2018sceneflowfields}}%
		\end{subfigure}
		\begin{subfigure}[c]{0.44\textwidth}
			\includegraphics[width=1\textwidth]{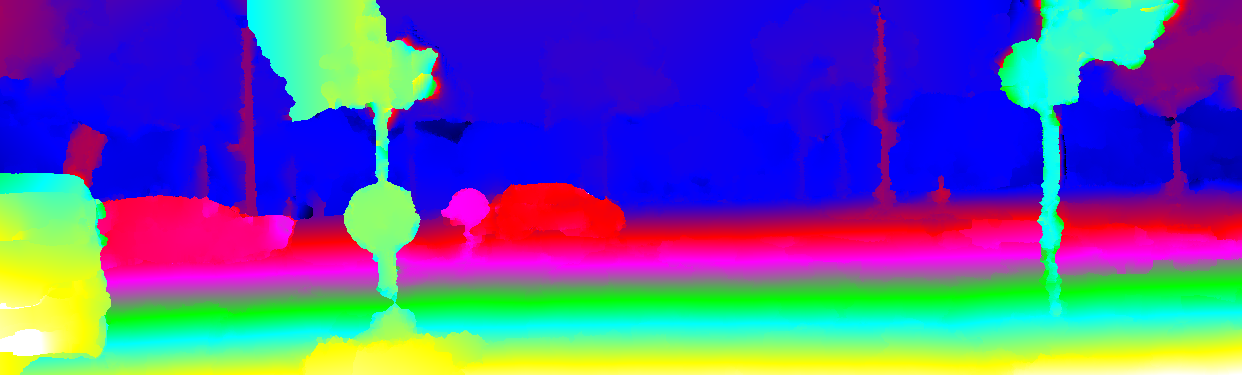}%
			\vspace{0.5mm}%
		\end{subfigure}
		\begin{subfigure}[c]{0.44\textwidth}
			\includegraphics[width=1\textwidth]{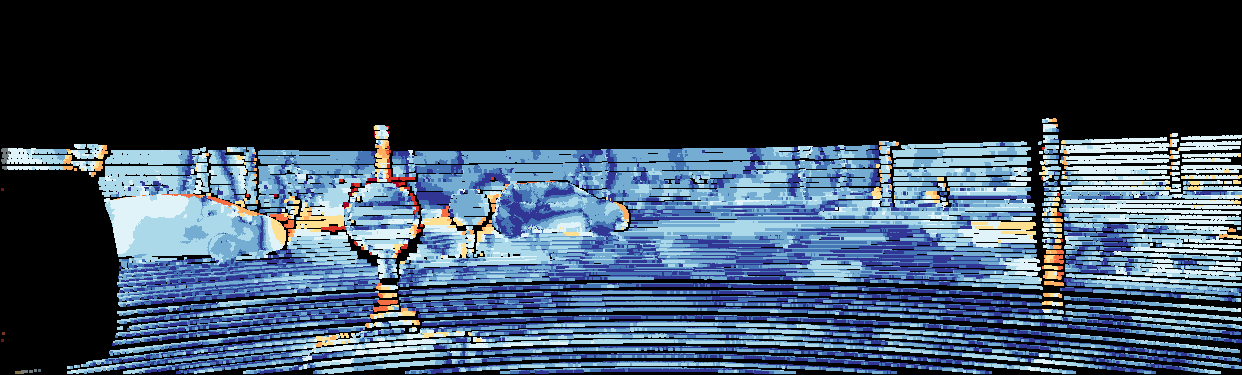}%
			\vspace{0.5mm}%
		\end{subfigure}
		\begin{subfigure}[c]{0.04\textwidth}
			\rotatebox[origin=c]{90}{D2}%
		\end{subfigure}
		\\%
		\hspace{0.04\textwidth}
		\begin{subfigure}[c]{0.44\textwidth}
			\includegraphics[width=1\textwidth]{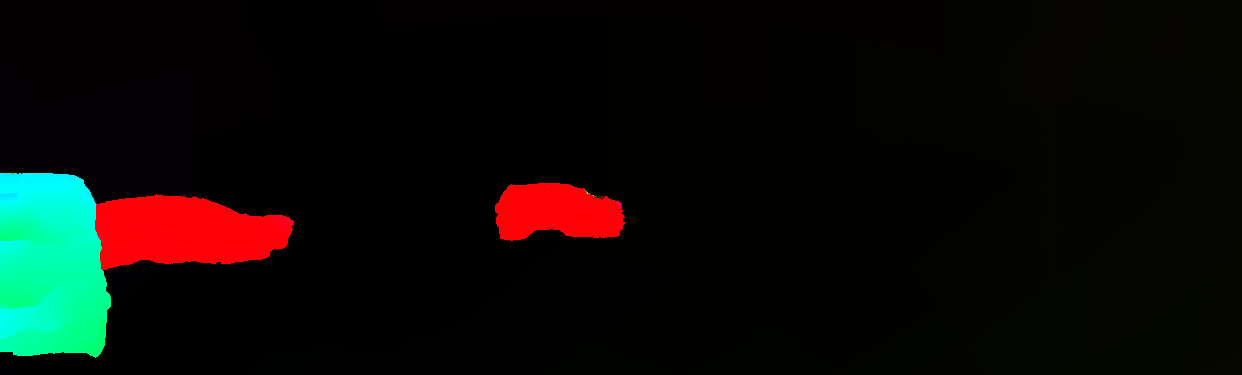}%
			\vspace{1mm}%
		\end{subfigure}
		\begin{subfigure}[c]{0.44\textwidth}
			\includegraphics[width=1\textwidth]{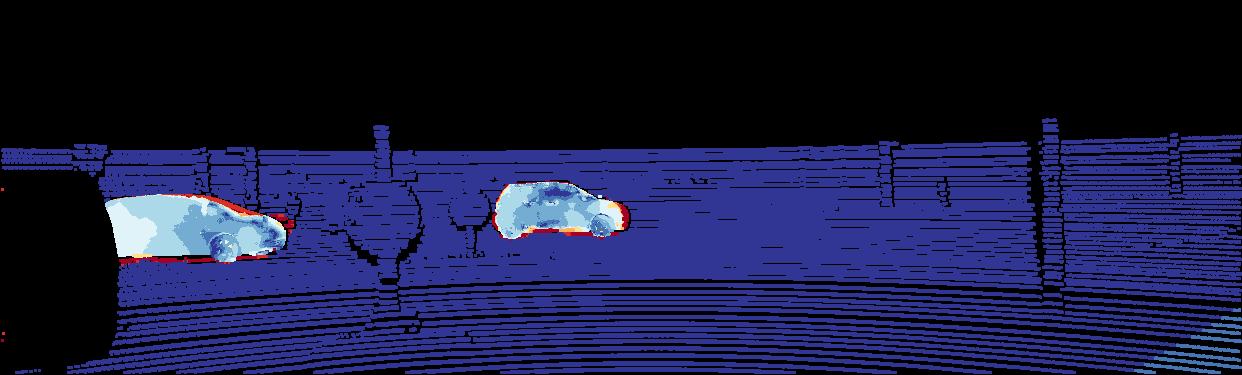}%
			\vspace{1mm}%
		\end{subfigure}
		\begin{subfigure}[c]{0.04\textwidth}
			\rotatebox[origin=c]{90}{Flow}%
		\end{subfigure}
		\\%
		\hspace{0.04\textwidth}
		\begin{subfigure}[c]{0.44\textwidth}
			\includegraphics[width=1\textwidth]{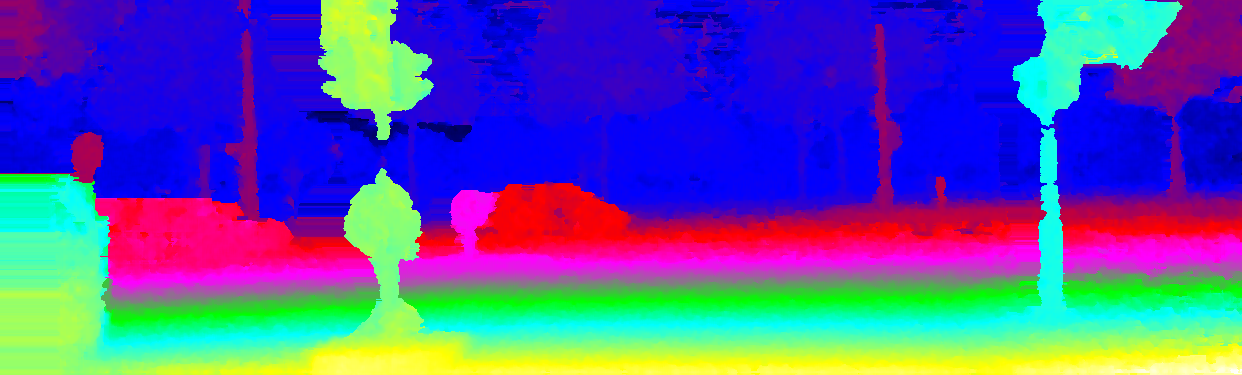}%
			\vspace{0.5mm}%
		\end{subfigure}
		\begin{subfigure}[c]{0.44\textwidth}
			\includegraphics[width=1\textwidth]{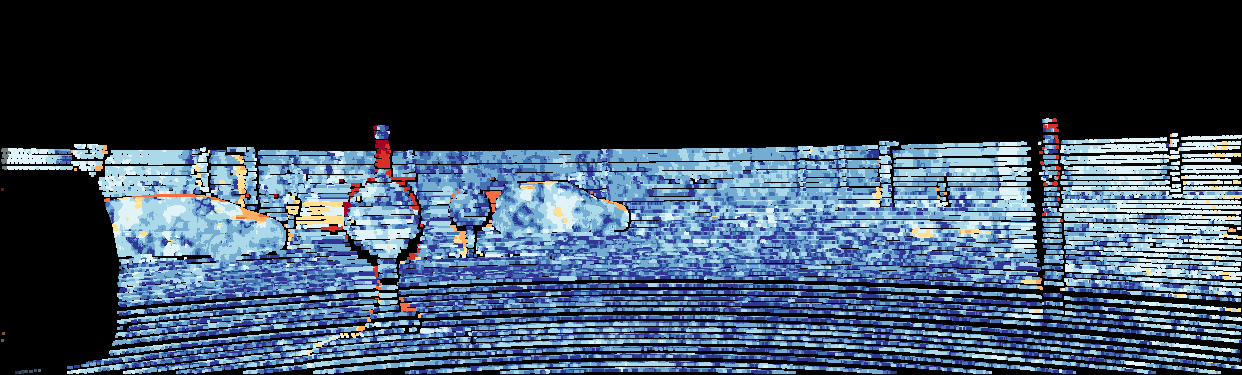}%
			\vspace{0.5mm}%
		\end{subfigure}
		\begin{subfigure}[c]{0.04\textwidth}
			\rotatebox[origin=c]{90}{D1}%
		\end{subfigure}
		\\%
		\begin{subfigure}[c][5mm][c]{0.04\textwidth}
			\rotatebox[origin=c]{90}{SGM+C+NL \cite{hirschmuller2008SGM,sun2014quantitative}}%
		\end{subfigure}
		\begin{subfigure}[c]{0.44\textwidth}
			\includegraphics[width=1\textwidth]{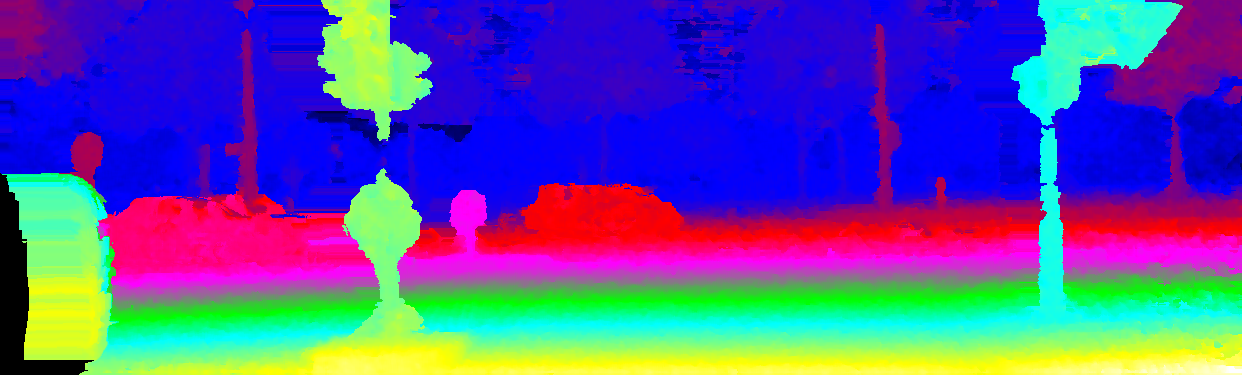}%
			\vspace{0.5mm}%
		\end{subfigure}
		\begin{subfigure}[c]{0.44\textwidth}
			\includegraphics[width=1\textwidth]{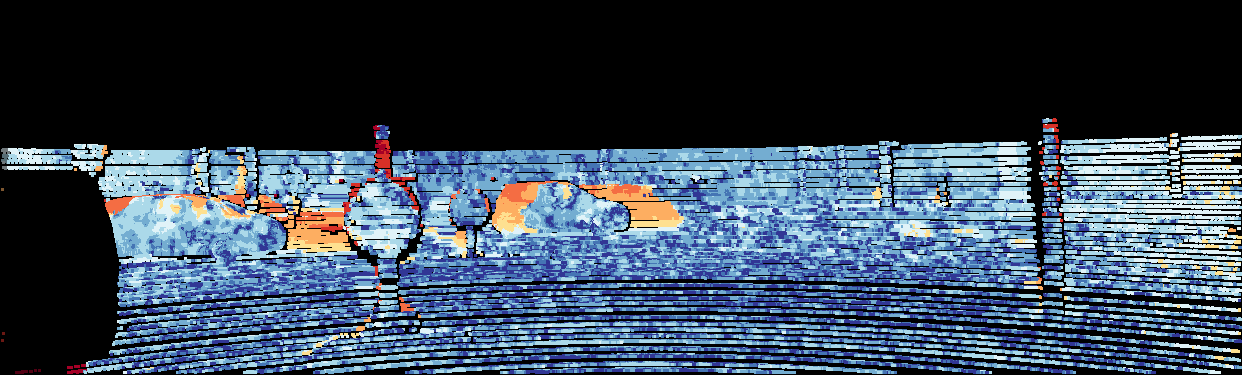}%
			\vspace{0.5mm}%
		\end{subfigure}
		\begin{subfigure}[c]{0.04\textwidth}
			\rotatebox[origin=c]{90}{D2}%
		\end{subfigure}
		\\%
		\hspace{0.04\textwidth}
		\begin{subfigure}[c]{0.44\textwidth}
			\includegraphics[width=1\textwidth]{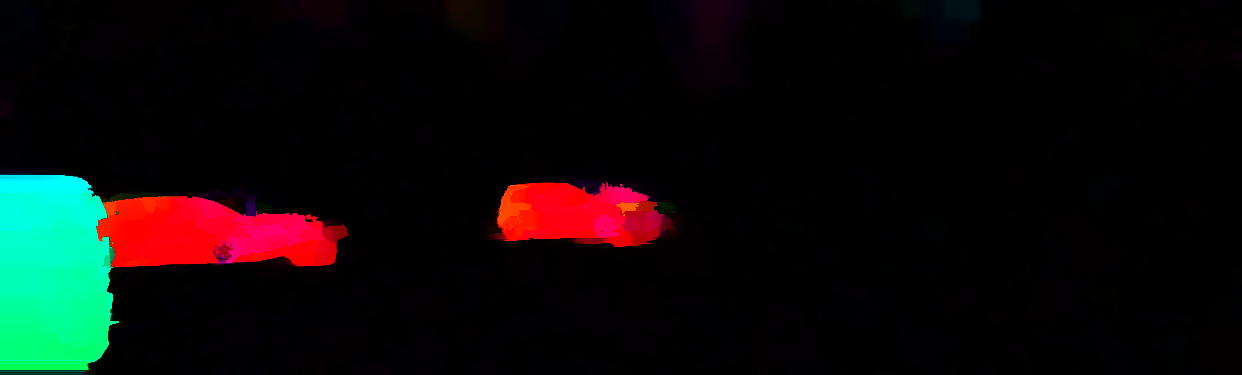}%
			\vspace{1mm}%
		\end{subfigure}
		\begin{subfigure}[c]{0.44\textwidth}
			\includegraphics[width=1\textwidth]{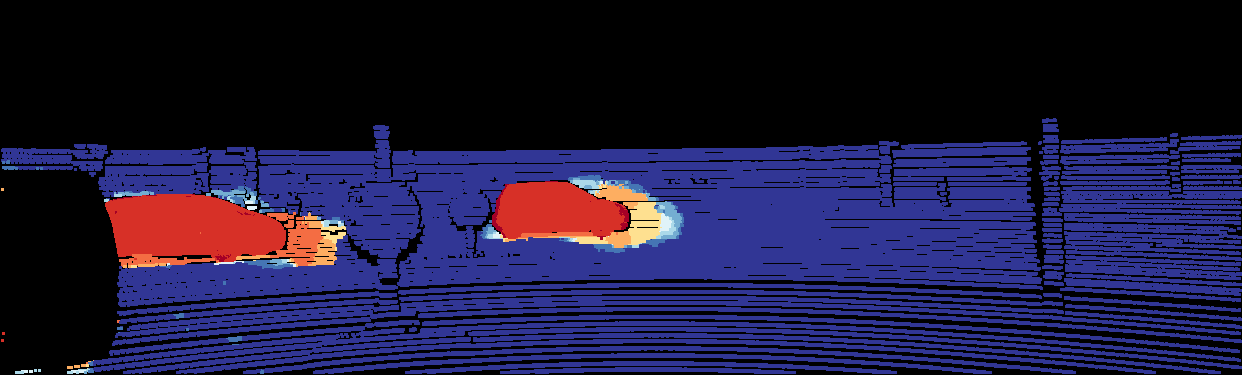}%
			\vspace{1mm}%
		\end{subfigure}
		\begin{subfigure}[c]{0.04\textwidth}
			\rotatebox[origin=c]{90}{Flow}%
		\end{subfigure}
		\\%
		\hspace{0.04\textwidth}
		\begin{subfigure}[c]{0.44\textwidth}
			\includegraphics[width=1\textwidth]{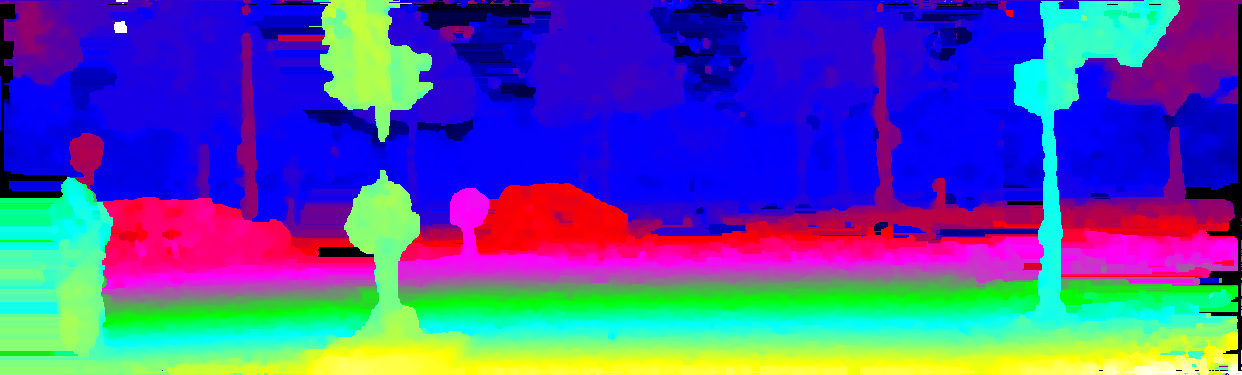}%
			\vspace{0.5mm}%
		\end{subfigure}
		\begin{subfigure}[c]{0.44\textwidth}
			\includegraphics[width=1\textwidth]{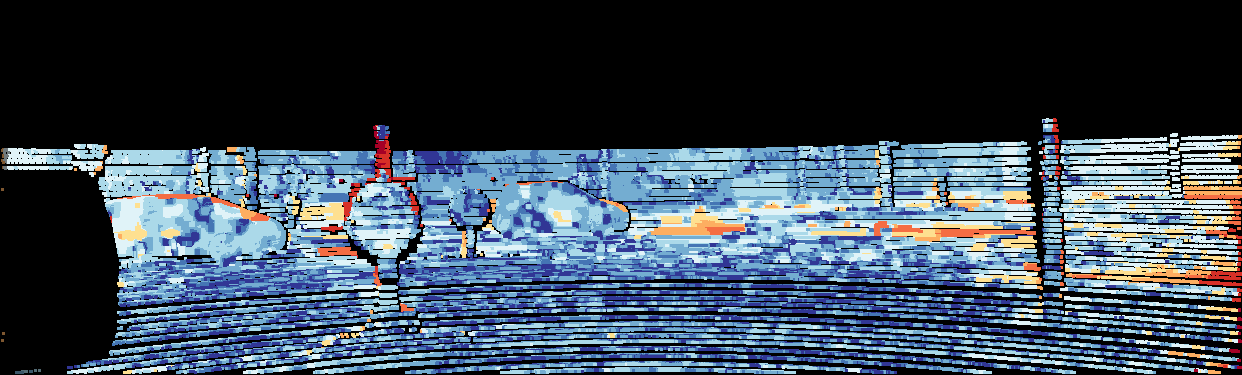}%
			\vspace{0.5mm}%
		\end{subfigure}
		\begin{subfigure}[c]{0.04\textwidth}
			\rotatebox[origin=c]{90}{D1}%
		\end{subfigure}
		\\%
		\begin{subfigure}[c][5mm][c]{0.04\textwidth}
			\rotatebox[origin=c]{90}{\bf Ours}%
		\end{subfigure}
		\begin{subfigure}[c]{0.44\textwidth}
			\includegraphics[width=1\textwidth]{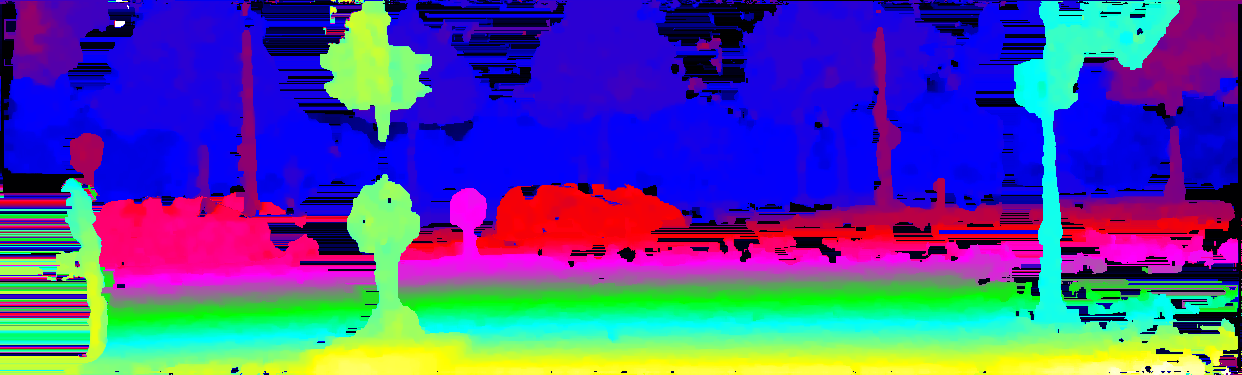}%
			\vspace{0.5mm}%
		\end{subfigure}
		\begin{subfigure}[c]{0.44\textwidth}
			\includegraphics[width=1\textwidth]{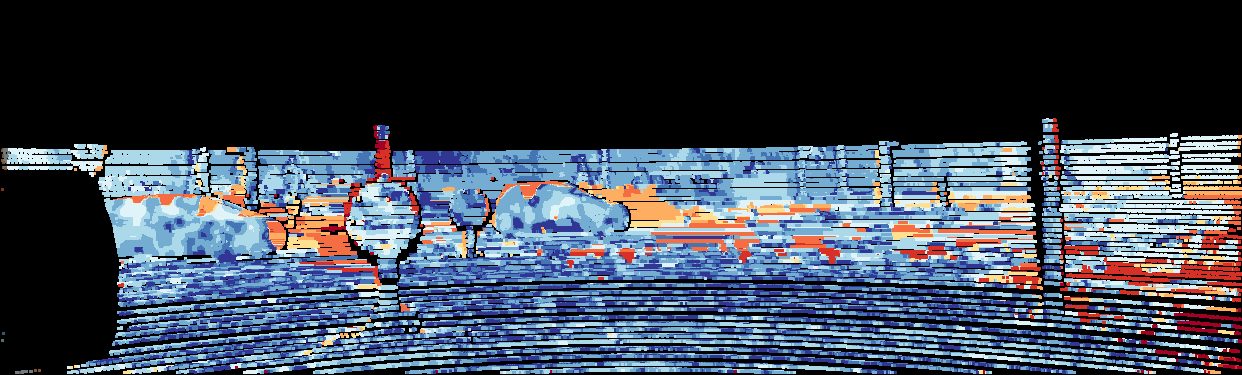}%
			\vspace{0.5mm}%
		\end{subfigure}
		\begin{subfigure}[c]{0.04\textwidth}
			\rotatebox[origin=c]{90}{D2}%
		\end{subfigure}
		\\%
		\hspace{0.04\textwidth}
		\begin{subfigure}[c]{0.44\textwidth}
			\includegraphics[width=1\textwidth]{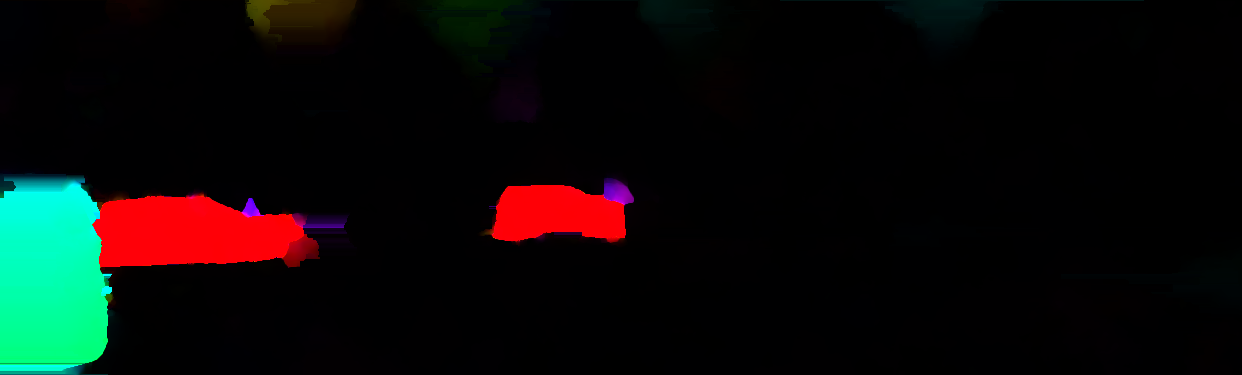}%
		\end{subfigure}
		\begin{subfigure}[c]{0.44\textwidth}
			\includegraphics[width=1\textwidth]{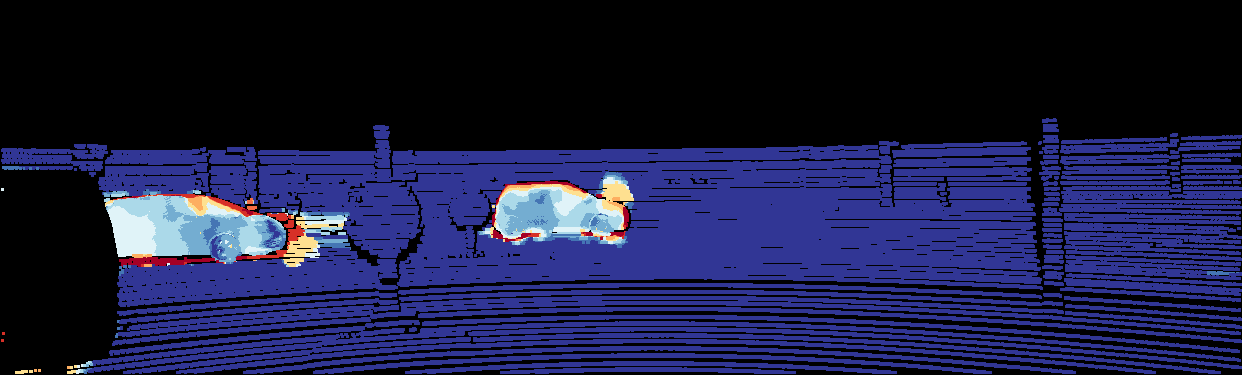}%
		\end{subfigure}
		\begin{subfigure}[c]{0.04\textwidth}
			\rotatebox[origin=c]{90}{Flow}%
		\end{subfigure}
	\end{center}
\caption{Visual comparison of the results on KITTI test image 8 for \mbox{SceneFlowFields} \cite{schuster2018sceneflowfields}, SGM+C+NL \cite{hirschmuller2008SGM,sun2014quantitative}, and our proposed method.}
\label{fig:comparison8}
\end{figure}

\begin{figure}[p]
	\begin{center}
		\begin{subfigure}[c]{1\textwidth}
			\centering
			\includegraphics[width=0.44\textwidth]{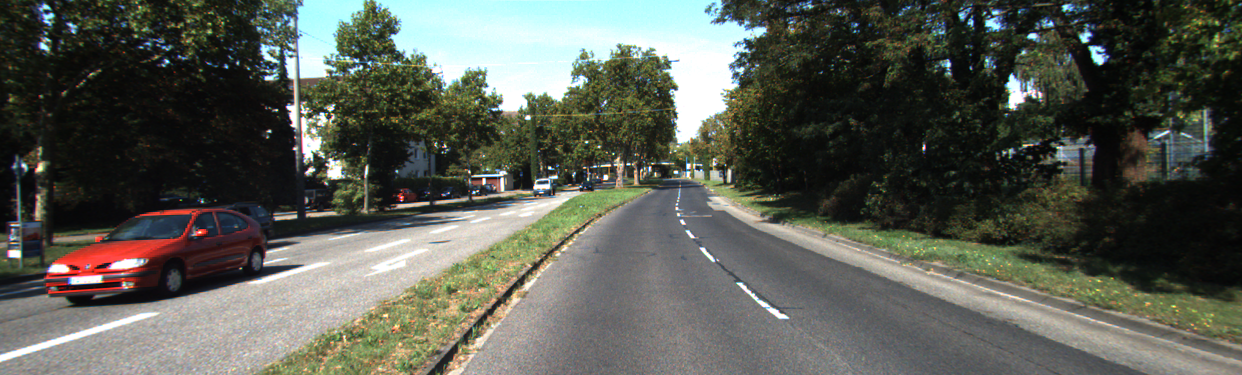}%
			\vspace{1mm}%
		\end{subfigure}
		\hfill
		\begin{subfigure}[c]{0.44\textwidth}
			\centering
			Estimates%
			\vspace{0.5mm}%
			\end{subfigure}
		\begin{subfigure}[c]{0.44\textwidth}
			\centering
			Error Maps%
			\vspace{0.5mm}%
		\end{subfigure}
		\hfill%
		\\%
		\hspace{0.04\textwidth}
		\begin{subfigure}[c]{0.44\textwidth}
			\includegraphics[width=1\textwidth]{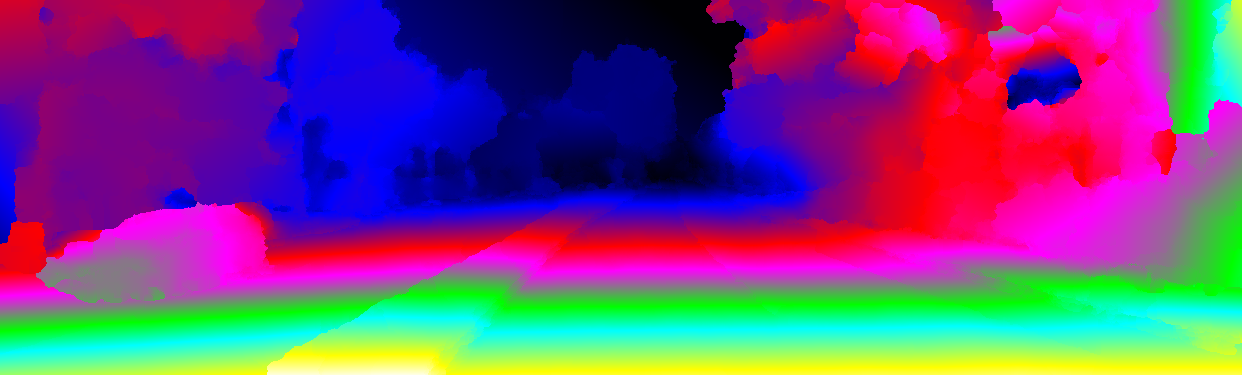}%
			\vspace{0.5mm}%
		\end{subfigure}
		\begin{subfigure}[c]{0.44\textwidth}
			\includegraphics[width=1\textwidth]{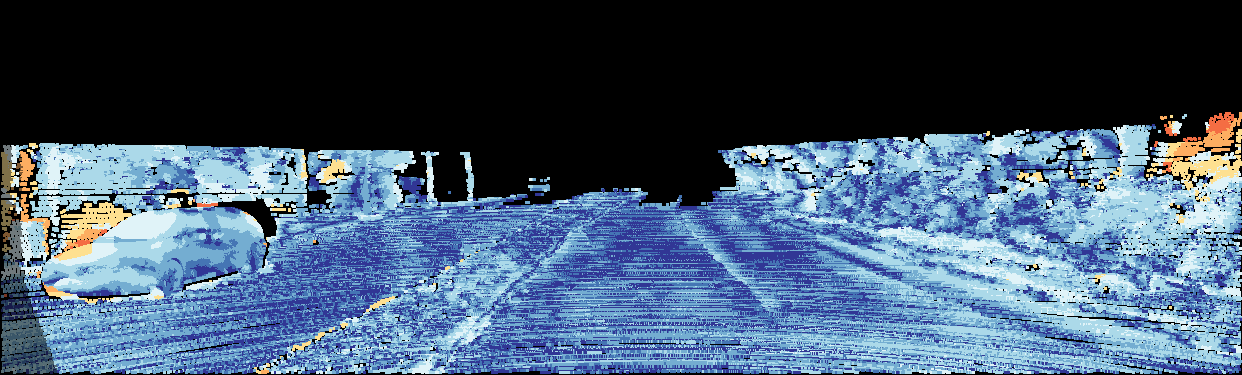}%
			\vspace{0.5mm}%
		\end{subfigure}
		\begin{subfigure}[c]{0.04\textwidth}
			\rotatebox[origin=c]{90}{D1}%
		\end{subfigure}
		\\%
		\begin{subfigure}[c][0.5mm][c]{0.04\textwidth}
			\rotatebox[origin=c]{90}{SceneFlowFields \cite{schuster2018sceneflowfields}}%
		\end{subfigure}
		\begin{subfigure}[c]{0.44\textwidth}
			\includegraphics[width=1\textwidth]{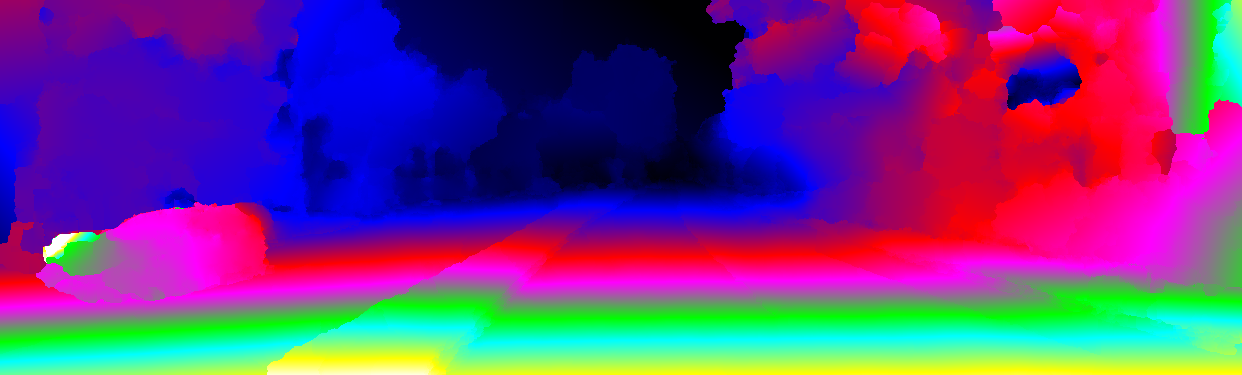}%
			\vspace{0.5mm}%
		\end{subfigure}
		\begin{subfigure}[c]{0.44\textwidth}
			\includegraphics[width=1\textwidth]{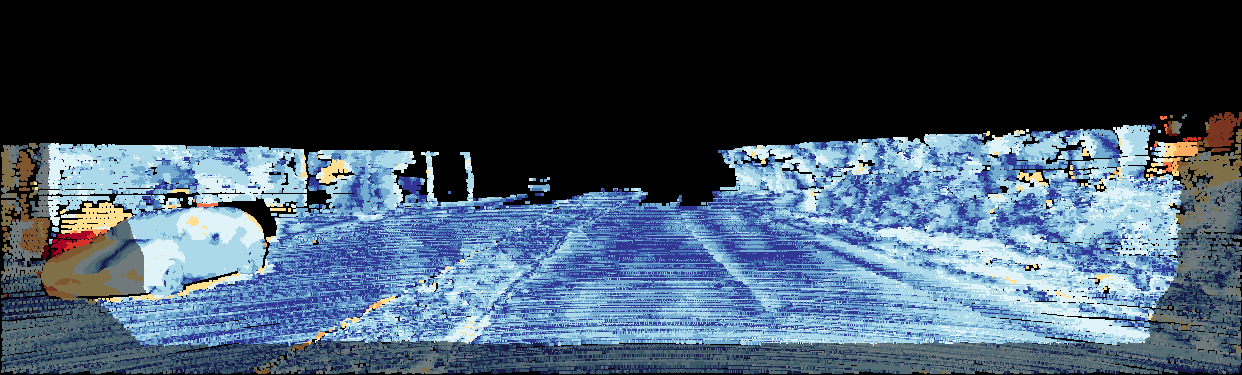}%
			\vspace{0.5mm}%
		\end{subfigure}
		\begin{subfigure}[c]{0.04\textwidth}
			\rotatebox[origin=c]{90}{D2}%
		\end{subfigure}
		\\%
		\hspace{0.04\textwidth}
		\begin{subfigure}[c]{0.44\textwidth}
			\includegraphics[width=1\textwidth]{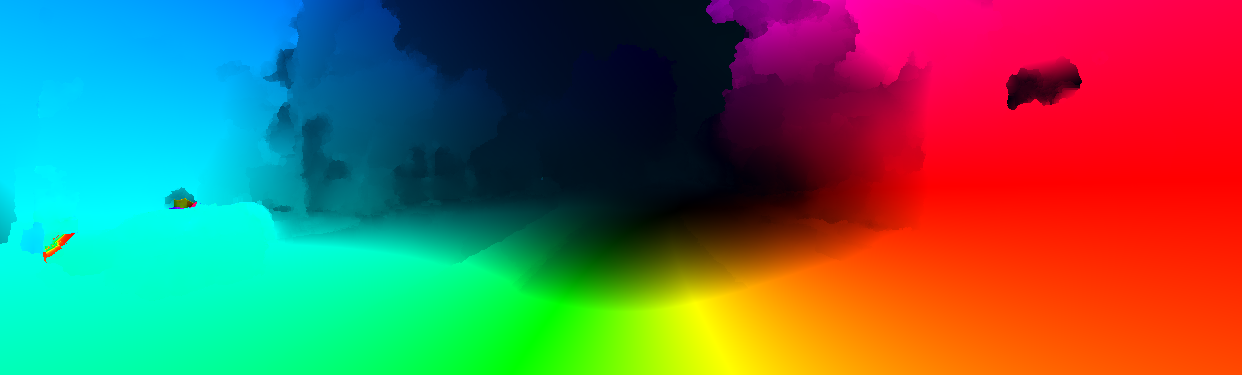}%
			\vspace{1mm}%
		\end{subfigure}
		\begin{subfigure}[c]{0.44\textwidth}
			\includegraphics[width=1\textwidth]{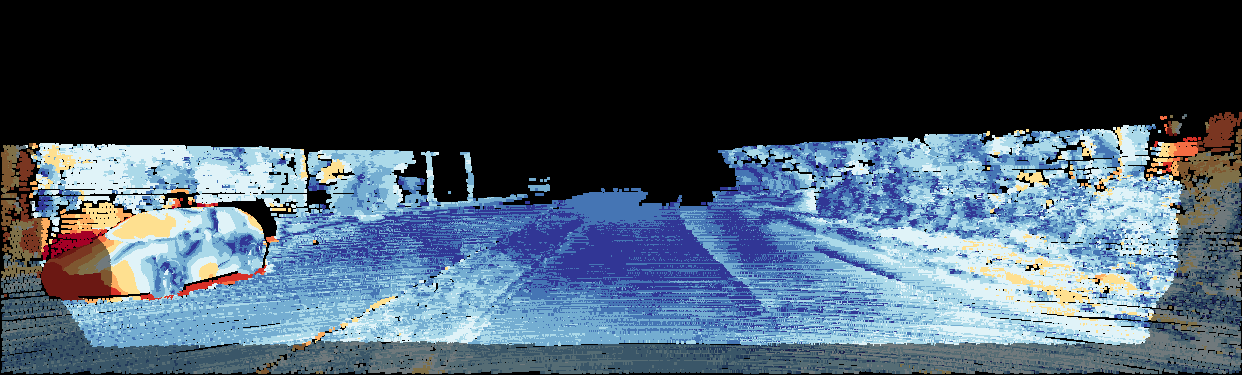}%
			\vspace{1mm}%
		\end{subfigure}
		\begin{subfigure}[c]{0.04\textwidth}
			\rotatebox[origin=c]{90}{Flow}%
		\end{subfigure}
		\\%
		\hspace{0.04\textwidth}
		\begin{subfigure}[c]{0.44\textwidth}
			\includegraphics[width=1\textwidth]{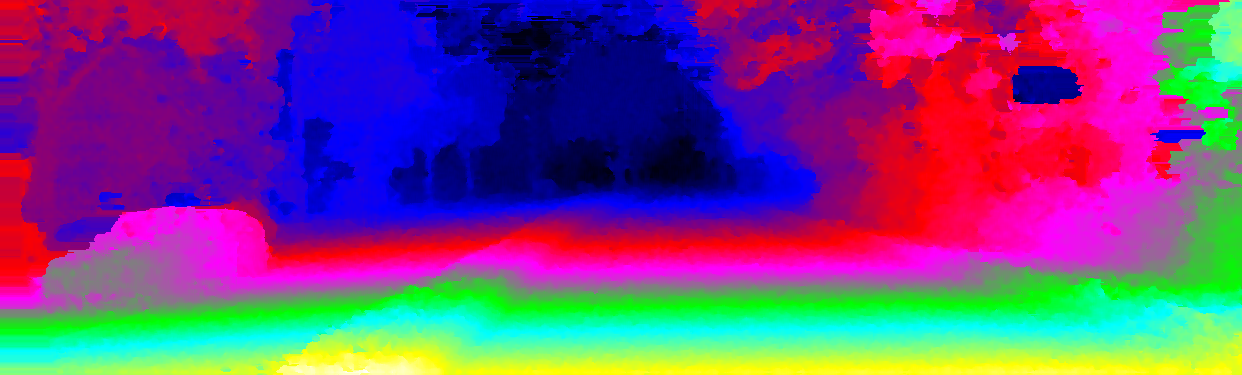}%
			\vspace{0.5mm}%
		\end{subfigure}
		\begin{subfigure}[c]{0.44\textwidth}
			\includegraphics[width=1\textwidth]{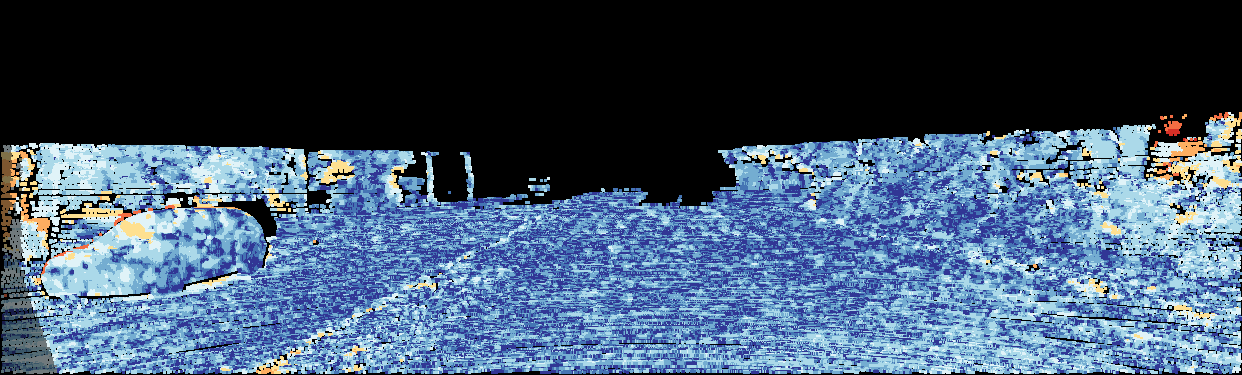}%
			\vspace{0.5mm}%
		\end{subfigure}
		\begin{subfigure}[c]{0.04\textwidth}
			\rotatebox[origin=c]{90}{D1}%
		\end{subfigure}
		\\%
		\begin{subfigure}[c][5mm][c]{0.04\textwidth}
			\rotatebox[origin=c]{90}{SGM+C+NL \cite{hirschmuller2008SGM,sun2014quantitative}}%
		\end{subfigure}
		\begin{subfigure}[c]{0.44\textwidth}
			\includegraphics[width=1\textwidth]{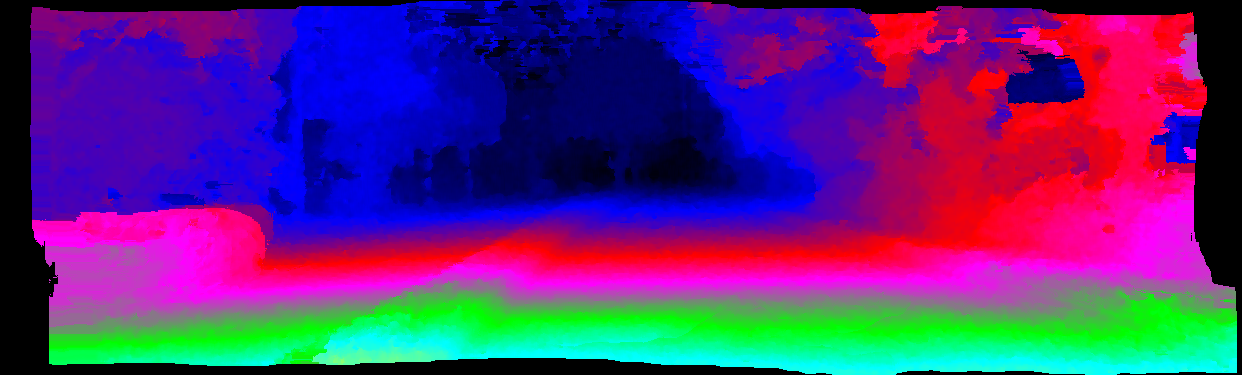}%
			\vspace{0.5mm}%
		\end{subfigure}
		\begin{subfigure}[c]{0.44\textwidth}
			\includegraphics[width=1\textwidth]{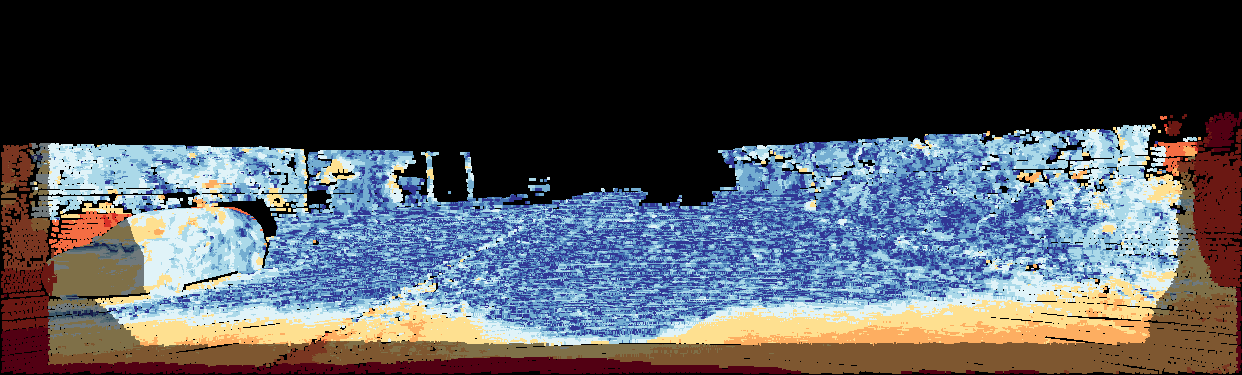}%
			\vspace{0.5mm}%
		\end{subfigure}
		\begin{subfigure}[c]{0.04\textwidth}
			\rotatebox[origin=c]{90}{D2}%
		\end{subfigure}
		\\%
		\hspace{0.04\textwidth}
		\begin{subfigure}[c]{0.44\textwidth}
			\includegraphics[width=1\textwidth]{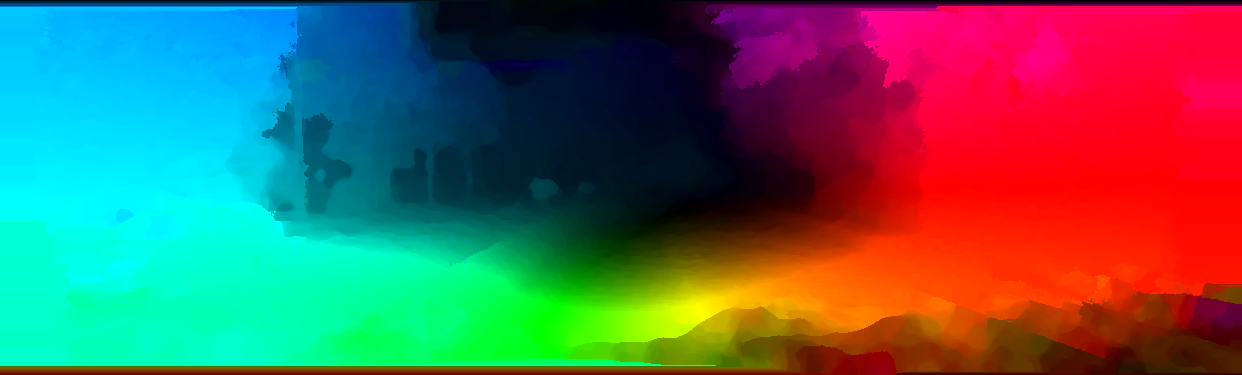}%
			\vspace{1mm}%
		\end{subfigure}
		\begin{subfigure}[c]{0.44\textwidth}
			\includegraphics[width=1\textwidth]{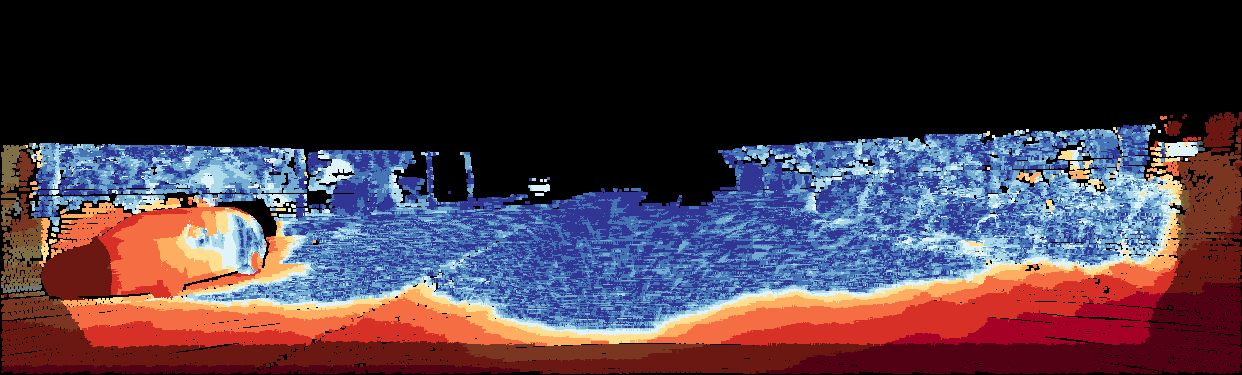}%
			\vspace{1mm}%
		\end{subfigure}
		\begin{subfigure}[c]{0.04\textwidth}
			\rotatebox[origin=c]{90}{Flow}%
		\end{subfigure}
		\\%
		\hspace{0.04\textwidth}
		\begin{subfigure}[c]{0.44\textwidth}
			\includegraphics[width=1\textwidth]{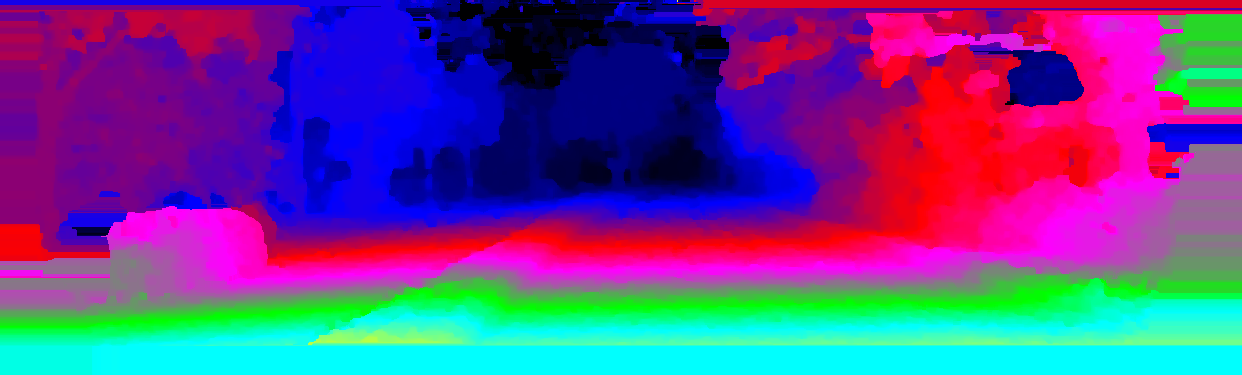}%
			\vspace{0.5mm}%
		\end{subfigure}
		\begin{subfigure}[c]{0.44\textwidth}
			\includegraphics[width=1\textwidth]{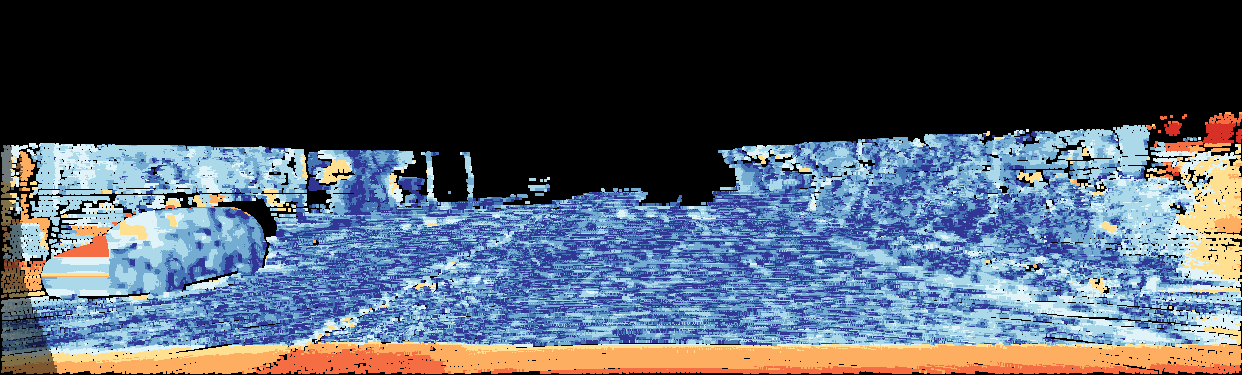}%
			\vspace{0.5mm}%
		\end{subfigure}
		\begin{subfigure}[c]{0.04\textwidth}
			\rotatebox[origin=c]{90}{D1}%
		\end{subfigure}
		\\%
		\begin{subfigure}[c][5mm][c]{0.04\textwidth}
			\rotatebox[origin=c]{90}{\bf Ours}%
		\end{subfigure}
		\begin{subfigure}[c]{0.44\textwidth}
			\includegraphics[width=1\textwidth]{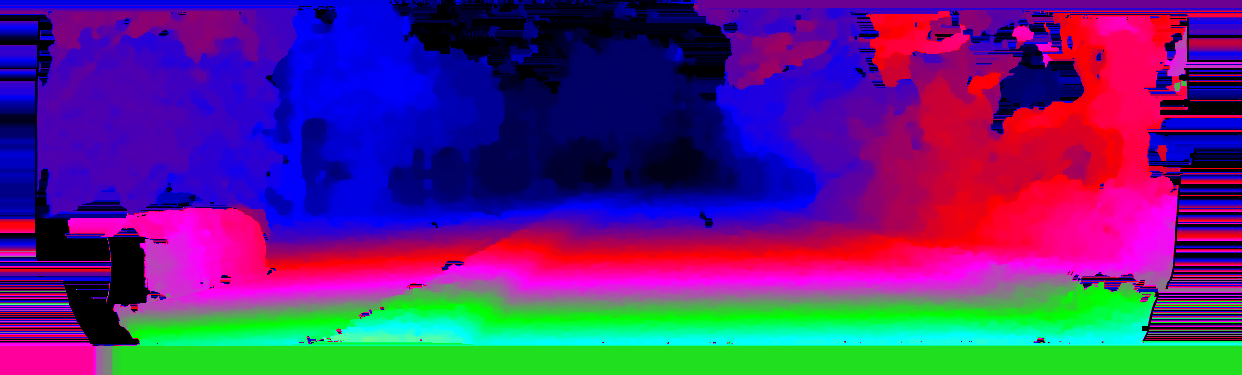}%
			\vspace{0.5mm}%
		\end{subfigure}
		\begin{subfigure}[c]{0.44\textwidth}
			\includegraphics[width=1\textwidth]{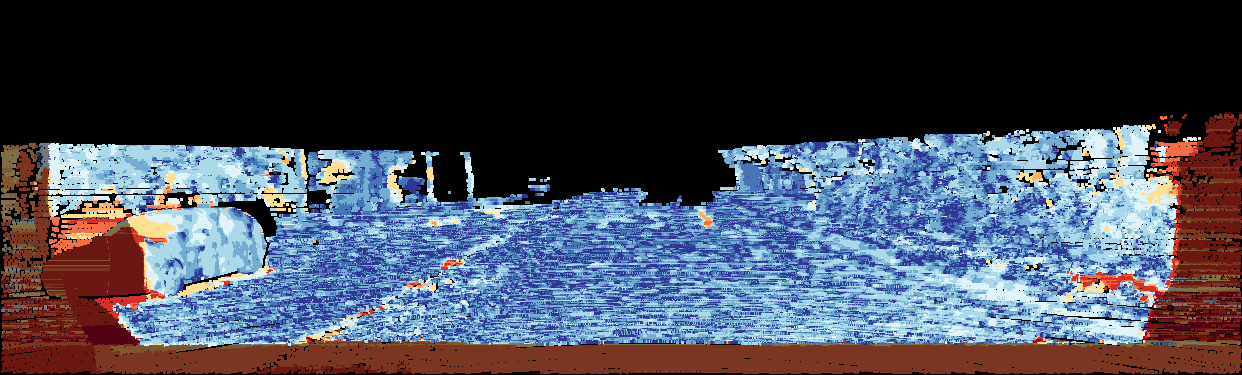}%
			\vspace{0.5mm}%
		\end{subfigure}
		\begin{subfigure}[c]{0.04\textwidth}
			\rotatebox[origin=c]{90}{D2}%
		\end{subfigure}
		\\%
		\hspace{0.04\textwidth}
		\begin{subfigure}[c]{0.44\textwidth}
			\includegraphics[width=1\textwidth]{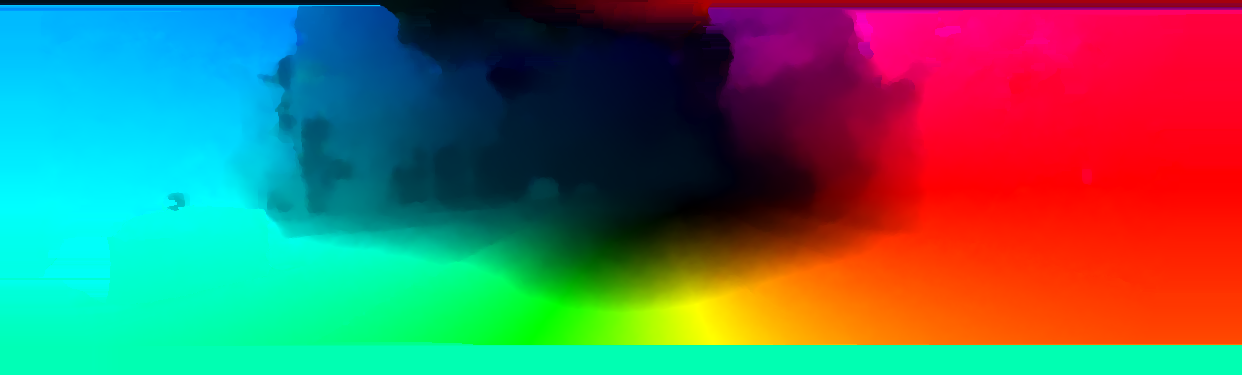}%
		\end{subfigure}
		\begin{subfigure}[c]{0.44\textwidth}
			\includegraphics[width=1\textwidth]{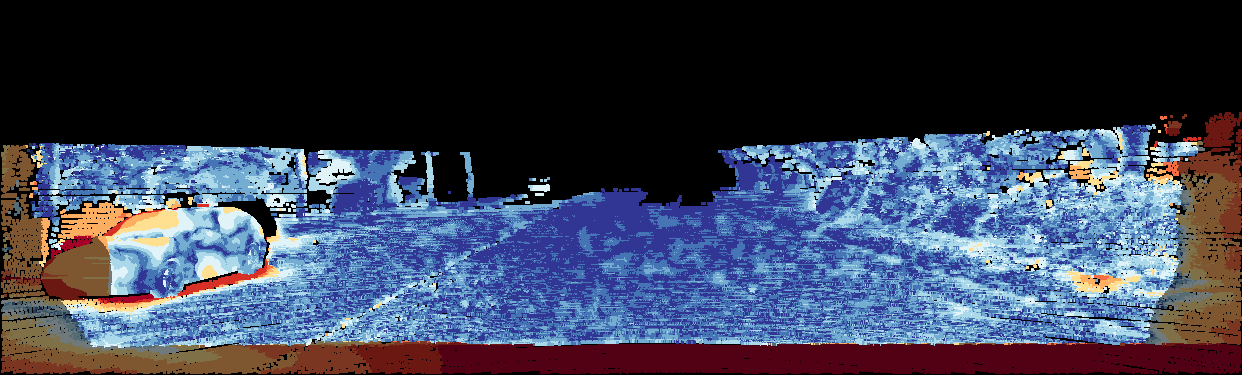}%
		\end{subfigure}
		\begin{subfigure}[c]{0.04\textwidth}
			\rotatebox[origin=c]{90}{Flow}%
		\end{subfigure}
	\end{center}
\caption{Visual comparison of the results on KITTI test image 13 for \mbox{SceneFlowFields} \cite{schuster2018sceneflowfields}, SGM+C+NL \cite{hirschmuller2008SGM,sun2014quantitative}, and our proposed method.}
\label{fig:comparison13}
\end{figure}

The most important part of our proposed method is that it extends 2D motion information to 3D space using depth. As described in \cref{fig:ofvssf}, this is a huge advantage. Compared to optical flow alone, scene flow additionally describes the motion in direction of viewing, providing full 3D motion information that many applications benefit from. 
We give visual examples of our results compared to other methods in \cref{fig:comparison8,fig:comparison13}. These figures show the color encoded two disparity maps and the optical flow along with the respective error maps. \cref{fig:comparison13} illustrates that most errors in our approach get introduced in areas where the motion leaves the image boundaries (shaded regions in the error maps) so that our recombination method can not reconstruct scene flow. Other methods that recombine stereo disparity and optical flow often fail even in the visible parts of a scene (see vehicles in \cref{fig:comparison8}).

Two major characteristics of our approach are discussed in more detail. First, as explained before, we compute non-dense scene flow. Secondly, the combination of stereo disparity and optical flow is fast compare to state-of-the-art scene flow algorithms.

\begin{table}[ht]
	\caption{Results of the public KITTI Scene Flow Benchmark \cite{menze2015object}. We compare all dual frame methods, i.e. methods that use only two consecutive frame pairs for computation. Results are given as average percentage of outliers according to the KITTI metric. \textit{Noc} shows the evaluation for non-occluded pixels only. \textit{Occ} gives results for all pixels. Our results are displayed for the original submitted scene flow (\textit{Est}) and for the automatic dense interpolation (\textit{All}) of the submission system.} \label{tab:kitti}
	\resizebox{\textwidth}{!}{
	\centering
	\begin{tabular}{c || r | r | r | r || r | r | r | r || r | r }
 & \multicolumn{4}{c||}{\textbf{Noc}} & \multicolumn{4}{c||}{\textbf{Occ}} & & \multicolumn{1}{|c}{\bf Run}\\
{\bf Method} & \multicolumn{1}{c|}{\bf D1} & \multicolumn{1}{c|}{\bf D2} & \multicolumn{1}{c|}{\bf Fl} & \multicolumn{1}{c||}{\bf SF} & \multicolumn{1}{c|}{\bf D1} & \multicolumn{1}{c|}{\bf D2} & \multicolumn{1}{c|}{\bf Fl} & \multicolumn{1}{c||}{\bf SF} & \multicolumn{1}{c|}{\bf Density} & \multicolumn{1}{c}{\bf time}\Bstrut\\ 
\hline
ISF \cite{behl2017bounding} & \textbf{4.02} & \textbf{4.69} & \textbf{4.69} & \textbf{6.45} & 4.46 & \textbf{5.95} & \textbf{6.22} & \textbf{8.08} & 100.00 \% & 600 s\Tstrut\\
SSF \cite{ren2017cascaded} & 4.03 & 5.99 & 5.40 & 8.25 & \textbf{4.42} & 7.02 & 7.14 & 10.07 & 100.00 \% & 300 s\\
OSF \cite{menze2015object} & 5.29 & 6.61 & 6.26 & 8.52 & 5.79 & 7.77 & 7.83 & 10.23 & 100.00 \% & 3000 s\\
CSF \cite{lv2016CSF} & 5.31 & 8.24 & 11.20 & 13.56 & 5.98 & 10.06 & 12.96 & 15.71 & 100.00 \% & 80 s\\
SceneFlowFields \cite{schuster2018sceneflowfields} & 6.08 & 8.11 & 10.39 & 12.99 & 6.57 & 10.69 & 12.88 & 15.78 & 100.00 \% & 65 s\\
PRSF \cite{vogel2013PRSF} & 5.84 & 8.10 & 9.36 & 11.53 & 6.24 & 12.69 & 13.83 & 16.44 & 100.00 \% & 150 s \\
{\bf Ours (Est)} & 4.61 & 8.23 & 9.47 & 16.43 & 4.68 & 10.61 & 11.84 & 19.81 & 81.24 \% & 29 s \\
SGM+SF \cite{hirschmuller2008SGM,hornacek2014sphereflow} & 6.31 & 10.20 & 14.89 & 17.86 & 6.84 & 15.60 & 21.67 & 24.98 & 100.00 \% & 2700 s \\
PCOF-LDOF \cite{derome2016prediction} & 8.02 & 11.28 & 13.80 & 19.68 & 8.46 & 20.99 & 18.33 & 29.27 & 100.00 \% & 50 s \\
{\bf Ours (All)} & 12.39 & 15.52 & 11.75 & 21.41 & 13.37 & 27.80 & 22.82 & 33.57 & 100.00 \% & 29 s \\
SGM+C+NL \cite{hirschmuller2008SGM,sun2014quantitative} & 6.31 & 16.63 & 25.84 & 29.84 & 6.84 & 28.25 & 35.61 & 40.33 & 100.00 \% & 270 s \\
SGM+LDOF \cite{hirschmuller2008SGM,brox2011large} & 6.31 & 17.36 & 29.87 & 33.64 & 6.84 & 28.56 & 39.33 & 43.67 & 100.00 \% & 83 s\\
DWBSF \cite{richardt2016dense} & 19.16 & 23.55 & 26.68 & 35.51 & 20.12 & 34.46 & 39.14 & 45.48 & 100.00 \% & 420 s \\
GCSF \cite{cech2011scene} & 13.72 & 23.63 & 38.05 & 45.21 & 14.21 & 33.41 & 46.40 & 53.54 & 100.00 \% & \textbf{2.4 s} \\
VSF \cite{huguet2007variational} & 25.31 & 50.24 & 41.28 & 60.78 & 26.38 & 57.08 & 49.28 & 66.90 & 100.00 \% & 7500 s \\
	\end{tabular}
	}
\end{table}

\subsection{Accuracy and Density} \label{sec:accuracy}
Sparsity of course is not a desired result, yet the non-dense nature of our method leads to accurate results. This can be seen in \cref{tab:kitti} where our sparse results (\textit{Est}) outperform other methods which combine stereo and optical flow as well as many of the dedicated scene flow algorithms. Our interpolated results (\textit{All}) are still better in comparison to SGM+C+NL \cite{hirschmuller2008SGM,sun2014quantitative} and SGM+LDOF \cite{hirschmuller2008SGM,brox2011large}, two methods that also combine depth and optical flow to obtain scene flow in a way similar to ours. This is an expected result because the optical flow algorithm we use is ranked higher in the respective KITTI benchmark \cite{geiger2012kitti}. However, interpolation of sparse scene flow as done by KITTI decreases the accuracy significantly. Other interpolation methods could lead to a higher accuracy for dense results \cite{schuster2018sceneflowfields}. Nevertheless, we outperform many of the dedicated scene flow algorithms like e.g. the variational approach of \cite{huguet2007variational}. In the end, the achieved density of about $81~\%$ (cf. \cref{tab:kitti}) is rather high considering that KITTIs data is recorded with a frame rate of $10$ which means that large parts of the visible scene leave the image boundaries at the next time step, even for slow ego-velocities. In KITTI, density is given by the amount of available ground truth pixels that are covered by the results. For the non-occluded areas, i.e. areas that are also visible in the next frame, we even achieve a density of $92.42~\%$.
Of course there exist methods that are ranked higher in the KITTI scene flow benchmark. Most of these methods make further assumptions on the observed scene which makes them less versatile. Furthermore, these methods solve scene flow estimation as a single task where geometry and 3D motion are estimated jointly. This allows for strong regularization mechanisms, e.g. the piece-wise rigid scene model that is used by \cite{vogel2013PRSF,menze2015object,lv2016CSF,behl2017bounding}, and has other advantages \cite{schuster2017towards}. But typically, the problem formulation in these methods results in a complex energy term that requires a lot of computational effort to minimize. This is reflected by the considerable long run times of the top performing methods on KITTI. 
We will investigate our comparable short run time in the next section.

\subsection{Run time} \label{runtime}
The second important aspect of our approach is the fast run time. We draw the following conclusion: The overall run time of any recombination method for scene flow is determined by the run time for depth and optical flow computation. In our case, the 29 seconds stem from 28 seconds computation time for Flow Fields+ and 1 second for disparity computation for both time steps using SGM. Combination time can be neglected. This means that sparse scene flow can be computed in real-time if real-time algorithms for the stereo and optical flow tasks are used. Even though the two subsidiary methods that we use are not the fastest in their respective field, our computation of scene flow from stereo and optical flow is at least two times faster than most methods and about one order of magnitude faster than the top performing method (cf. \cref{tab:kitti}).

\section{Conclusion} \label{sec:conclusion}
In summary, we have presented a straightforward approach to compute scene flow. Accuracy and run time only depend on the algorithms that are used to compute depth and optical flow. Thus, scene flow estimation in real-time is possible in theory. We have demonstrated that even the basic combination of optical flow and disparity leads to state-of-the-art results which outperform other combination methods and many other dedicated scene flow algorithms. Most of the remaining errors are due to out-of-bounds motions. We have explained why scene flow is preferable to optical flow supported by a visual example.

For future work, we would like to improve the overall results by replacing the automatic interpolation of the KITTI submission system by a more appropriate method like e.g. the one that is used in \cite{schuster2018sceneflowfields}. It is also imaginable to extend our approach to more than two frame pairs and to apply a intermediate validation strategy like in \cite{wasenmueller2014correspondence}.

\newpage
\bibliography{bib}

\begin{thebibliography}{10}

\bibitem{menze2015object}
Menze, M., Geiger, A.:
\newblock Object scene flow for autonomous vehicles.
\newblock In: Conference on Computer Vision and Pattern Recognition (CVPR).
  (2015)

\bibitem{vedula1999three}
Vedula, S., Baker, S., Rander, P., Collins, R., Kanade, T.:
\newblock Three-dimensional scene flow.
\newblock In: International Conference on Computer Vision (ICCV). (1999)

\bibitem{huguet2007variational}
Huguet, F., Devernay, F.:
\newblock A variational method for scene flow estimation from stereo sequences.
\newblock In: International Conference on Computer Vision (ICCV). (2007)

\bibitem{basha2013multi}
Basha, T., Moses, Y., Kiryati, N.:
\newblock Multi-view scene flow estimation: {A} view centered variational
  approach.
\newblock International Journal of Computer Vision (IJCV) (2013)

\bibitem{vogel2013PRSF}
Vogel, C., Schindler, K., Roth, S.:
\newblock Piecewise rigid scene flow.
\newblock In: International Conference on Computer Vision (ICCV). (2013)

\bibitem{lv2016CSF}
Lv, Z., Beall, C., Alcantarilla, P.F., Li, F., Kira, Z., Dellaert, F.:
\newblock A continuous optimization approach for efficient and accurate scene
  flow.
\newblock In: European Conference on Computer Vision (ECCV). (2016)

\bibitem{neoral2017object}
Neoral, M., Šochman, J.:
\newblock Object scene flow with temporal consistency.
\newblock In: Computer Vision Winter Workshop (CVWW). (2017)

\bibitem{vogel2015PRSM}
Vogel, C., Schindler, K., Roth, S.:
\newblock {3D} scene flow estimation with a piecewise rigid scene model.
\newblock International Journal of Computer Vision (IJCV) (2015)

\bibitem{taniai2017fsf}
Taniai, T., Sinha, S.N., Sato, Y.:
\newblock Fast multi-frame stereo scene flow with motion segmentation.
\newblock In: {Conference on Computer Vision and Pattern Recognition (CVPR)}.
  (2017)

\bibitem{schuster2018sceneflowfields}
Schuster, R., Wasenmüller, O., Kuschk, G., Bailer, C., Stricker, D.:
\newblock {SceneFlowFields}: {D}ense interpolation of sparse scene flow
  correspondences.
\newblock In: Winter Conference on Applications of Computer Vision (WACV).
  (2018)

\bibitem{hornacek2014sphereflow}
Hornacek, M., Fitzgibbon, A., Rother, C.:
\newblock {SphereFlow}: 6 {DoF} scene flow from {RGB-D} pairs.
\newblock In: Conference on Computer Vision and Pattern Recognition (CVPR).
  (2014)

\bibitem{jaimez2015primal}
Jaimez, M., Souiai, M., Gonzalez-Jimenez, J., Cremers, D.:
\newblock A primal-dual framework for real-time dense {RGB-D} scene flow.
\newblock In: International Conference on Robotics and Automation (ICRA).
  (2015)

\bibitem{yoshida2017time}
Yoshida, T., Wasenm\"{u}ller, O., Stricker, D.:
\newblock Time-of-flight sensor depth enhancement for automotive exhaust gas.
\newblock In: International Conference on Image Processing (ICIP). (2017)

\bibitem{fortun2015optical}
Fortun, D., Bouthemy, P., Kervrann, C.:
\newblock Optical flow modeling and computation: {A} survey.
\newblock Computer Vision and Image Understanding (CVIU) (2015)

\bibitem{hirschmuller2008SGM}
Hirschmuller, H.:
\newblock Stereo processing by semiglobal matching and mutual information.
\newblock Transactions on Pattern Analysis and Machine Intelligence (PAMI)
  (2008)

\bibitem{yamaguchi2014efficient}
Yamaguchi, K., McAllester, D., Urtasun, R.:
\newblock Efficient joint segmentation, occlusion labeling, stereo and flow
  estimation.
\newblock In: European Conference on Computer Vision (ECCV). (2014)

\bibitem{bailer2015flow}
Bailer, C., Taetz, B., Stricker, D.:
\newblock Flow {F}ields: Dense correspondence fields for highly accurate large
  displacement optical flow estimation.
\newblock In: International Conference on Computer Vision (ICCV). (2015)

\bibitem{bailer2017cnn}
Bailer, C., Varanasi, K., Stricker, D.:
\newblock {CNN}-based patch matching for optical flow with thresholded hinge
  embedding loss.
\newblock In: Conference on Computer Vision and Pattern Recognition (CVPR).
  (2017)

\bibitem{bailer2017optical}
Bailer, C., Taetz, B., Stricker, D.:
\newblock Optical {F}low {F}ields: {D}ense correspondence fields for highly
  accurate large displacement optical flow estimation.
\newblock arXiv preprint arXiv:1703.02563 (2017)

\bibitem{revaud2015epic}
Revaud, J., Weinzaepfel, P., Harchaoui, Z., Schmid, C.:
\newblock Epic{F}low: Edge-preserving interpolation of correspondences for
  optical flow.
\newblock In: Conference on Computer Vision and Pattern Recognition (CVPR).
  (2015)

\bibitem{hel2005real}
Hel-Or, Y., Hel-Or, H.:
\newblock Real-time pattern matching using projection kernels.
\newblock Transactions on Pattern Analysis and Machine Intelligence (PAMI)
  (2005)

\bibitem{behl2017bounding}
Behl, A., Jafari, O.H., Mustikovela, S.K., Alhaija, H.A., Rother, C., Geiger,
  A.:
\newblock Bounding boxes, segmentations and object coordinates: {H}ow important
  is recognition for 3d scene flow estimation in autonomous driving scenarios?
\newblock In: International Conference on Computer Vision (ICCV). (2017)

\bibitem{ren2017cascaded}
Ren, Z., Sun, D., Kautz, J., Sudderth, E.B.:
\newblock Cascaded scene flow prediction using semantic segmentation.
\newblock In: International Conference on 3DVision (3DV). (2017)

\bibitem{derome2016prediction}
Derome, M., Plyer, A., Sanfourche, M., Le~Besnerais, G.:
\newblock A prediction-correction approach for real-time optical flow
  computation using stereo.
\newblock In: German Conference on Pattern Recognition (GCPR). (2016)

\bibitem{sun2014quantitative}
Sun, D., Roth, S., Black, M.J.:
\newblock A quantitative analysis of current practices in optical flow
  estimation and the principles behind them.
\newblock International Journal of Computer Vision (IJCV) (2014)

\bibitem{brox2011large}
Brox, T., Malik, J.:
\newblock Large displacement optical flow: {D}escriptor matching in variational
  motion estimation.
\newblock Transactions on Pattern Analysis and Machine Intelligence (PAMI)
  (2011)

\bibitem{richardt2016dense}
Richardt, C., Kim, H., Valgaerts, L., Theobalt, C.:
\newblock Dense wide-baseline scene flow from two handheld video cameras.
\newblock In: International Conference on 3D Vision (3DV). (2016)

\bibitem{cech2011scene}
{\v{C}}ech, J., Sanchez-Riera, J., Horaud, R.:
\newblock Scene flow estimation by growing correspondence seeds.
\newblock In: Conference on Computer Vision and Pattern Recognition (CVPR).
  (2011)

\bibitem{geiger2012kitti}
Geiger, A., Lenz, P., Urtasun, R.:
\newblock Are we ready for autonomous driving? {T}he {KITTI} vision benchmark
  suite.
\newblock In: Conference on Computer Vision and Pattern Recognition (CVPR).
  (2012)

\bibitem{schuster2017towards}
Schuster, R., Wasenm\"uller, O., Kuschk, G., Bailer, C., Stricker, D.:
\newblock Towards flow estimation in automotive scenarios.
\newblock In: ACM Computer Science in Cars Symposium (CSCS). (2017)

\bibitem{wasenmueller2014correspondence}
Wasenm\"{u}ller, O., Krolla, B., Michielin, F., Stricker, D.:
\newblock Correspondence chaining for enhanced dense {3D} reconstruction.
\newblock International Conference on Computer Graphics, Visualization and
  Computer Vision (WSCG) (2014)

\end{thebibliography}

\end{document}